\documentclass[letterpaper, 10 pt, conference]{ieeeconf}  

\IEEEoverridecommandlockouts                              

\usepackage{amsmath} 
\usepackage{amssymb}  
\usepackage{graphicx,dblfloatfix}
\usepackage{epstopdf}
\usepackage{booktabs}
\usepackage{mathtools}
\usepackage{bm}
\usepackage{tabularx}
\usepackage[labelfont=bf]{caption,subcaption}
\usepackage{cite}
\usepackage{siunitx}
\usepackage{gensymb}
\usepackage[linesnumbered, ruled, vlined, english]{algorithm2e}
\usepackage[super]{nth}
\usepackage{xcolor}

\SetCommentSty{mycommfont}

\title{\LARGE \bf
Dexterous Manipulation Graphs}

\author{Silvia Cruciani$^{1}$, Christian Smith$^{1}$, Danica Kragic$^{1}$ and Kaiyu Hang$^{2}$
\thanks{$^{1}$Silvia Cruciani, Christian Smith and Danica Kragic are with the Robotics, Perception and Learning Lab, EECS at KTH Royal Institute of Technology, Stockholm, Sweden.
{\tt\small\{cruciani, ccs, dani\}@kth.se}}%
\thanks{$^{2}$Kaiyu Hang is with the Robotics Institute and Institute for Advanced Study at  Hong Kong University of Science and Technology, Hong Kong.
{\tt\small kaiyu@hust.hk}}%
\thanks{$^*$This work was supported by the European Union framework program H2020-645403 RobDREAM.}
}

\begin{document}
	
\maketitle
\thispagestyle{empty}
\pagestyle{empty}

\begin{abstract}
	
We propose the Dexterous Manipulation Graph as a tool to address in-hand manipulation and reposition an object inside a robot's end-effector. This graph is used to plan a sequence of manipulation primitives so to bring the object to the desired end pose. This sequence of primitives is translated into motions of the robot to move the object held by the end-effector. We use a dual arm robot with parallel grippers to test our method on a real system and show successful planning and execution of in-hand manipulation. 
\end{abstract}

\section{Introduction} \label{sec_introduction}
Repositioning an object inside a robot's hand, known as in-hand manipulation, is an open problem in robotics. Human in-hand manipulation is a rich combination of various grasping movements, including regrasping, sliding and rotating the object. These motions are enabled by the complex human hands, and in particular by their dexterity. 

The dexterity of a hand is the ability to perform motions that manipulate a grasped object. In robotics, two terms are commonly used:
\begin{enumerate}
	\item \emph{Intrinsic dexterity} denotes the ability of the robot's hand to manipulate the object using its several degrees of freedom (DoF). End-effectors with high intrinsic dexterity often mimic the structure of the human hand\cite{ozawa, bicchi}. Alternatively, the hand can be simpler and the end-effector is designed specifically for the particular task it has to solve \cite{rahman, bircher_2}.
	\item \emph{Extrinsic dexterity} is the ability to compensate for the lack of DoF using external supports, such as friction, gravity and contact surfaces \cite{chavan-dafle_2}. It enables dexterous manipulation also with simple parallel grippers.
\end{enumerate}

In this paper, we use a dual arm robot to control the pose of an object held by the robot. Dual arm robots are becoming widely diffused (e.g. Baxter, Yumi), but the most common end-effector is still a simple parallel gripper. We enhance the lack of DoF of the gripper by using the two arms. Therefore, our system combines intrinsic and extrinsic dexterity: while it uses a poorly dexterous end-effector, it compensates with the several DoF available in a dual arm system. These DoF are external to the gripper alone, but cannot be seen entirely as external to the system since they can be fully controlled. 

The main focus of dexterous manipulation is to move an object held by the gripper to achieve different grasp poses. Past works mostly focused on the end-effector design and control. Therefore, there is a lack of quick and intuitive solutions for planning the in-hand motion of the object.

\begin{figure}[t]
	\centering
	\includegraphics[width=0.5\textwidth]{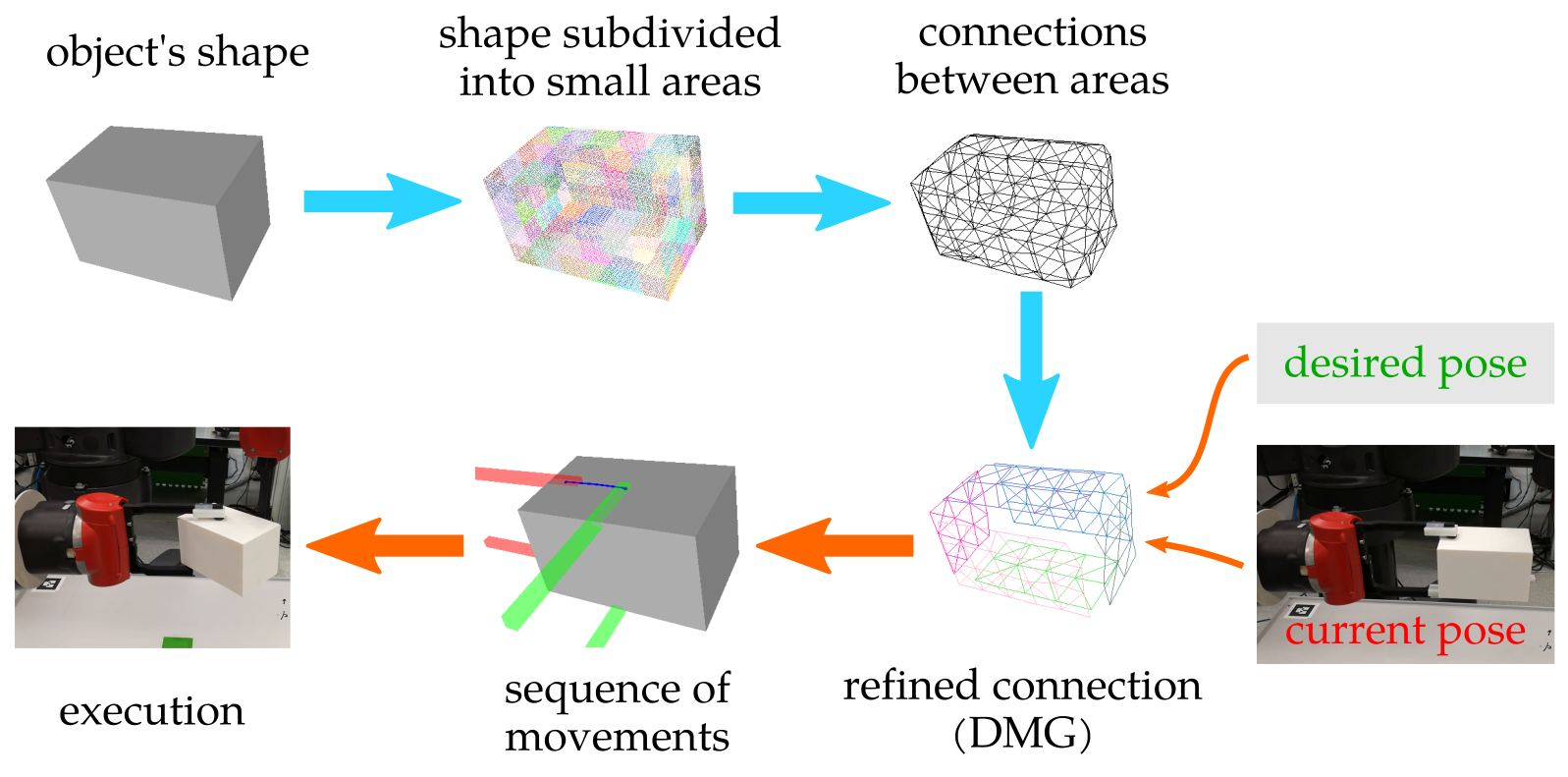}
	\caption{The blue arrows show the DMG's derivation: the object's shape is subdivided into small areas, and their connectivity is refined to describe the admissible motions of one fingertip on the object. The orange arrows show the progression of the repositioning: the DMG provides a sequence of movements from the initial fingers'contacts (in red) to the desired ones (in green); then, this sequence is executed by the robot.}\label{fig_visual_abstract}
\end{figure}

We propose the Dexterous Manipulation Graph (DMG) as a solution for planning dexterous manipulation. The DMG is a graph representation of the possible motions of one finger in contact with the object's surface. This graph is used to plan sequences of movements, which we call manipulation primitives, to change the contact points on the object so that it is manipulated towards the desired pose. This sequence of movements is then translated into motions of the dual arm robot. Fig.~\ref{fig_visual_abstract} summarizes the procedure for generating the DMG, and shows a simple example of its use for in-hand repositioning. 

In contrast with previous works, we do not require an accurate contact modeling. Our solution provides a way to move objects inside the gripper, even with complex object shapes and for large scale manipulation.

\section{Related Work} \label{sec_related_work}
Dual arm systems can be used to manipulate objects using two grippers at the same time \cite{smith}. In particular, we use the Extended Cooperative Task Space (ECTS) \cite{park_1} to formulate the execution of the motions necessary to reposition an object to the target pose. The ECTS task specification \cite{park_2} describes different manipulation scenarios, but it does not address the problem of moving an object inside the robot's gripper.

The method for dual arm manipulation proposed in \cite{murooka} exploits a state transition graph of object motions. This graph is used to achieve whole-body contact using a humanoid robot, so it exploits both contacts with the hands and with the forearms and possibly other links. In this case, the transition graph is generated by testing the outcome of basic operations on the object, e.g. lifting or sliding. Therefore, this method requires a physical model of both the object and the environment. Moreover, it does not deal with complex object shapes because the purpose is changing the configuration of big objects for stable transportation.

The ``IK-switch" move designed in \cite{xian} allows a system of two manipulators to reorient a grasped object. However, it is a regrasping move, therefore the change in configuration is obtained by releasing and grasping the object multiple times, and not by moving the contact points on it.

Despite the use of a dual arm system, our main objective is dexterous manipulation with the end-effector. In-hand manipulation is a widely studied problem. However, most of the available works focus on the control aspect of this problem and often rely on high fidelity dynamic models, and on highly precise and rich sensory input from motion capture systems and force and tactile sensors.

The HFTS grasping method \cite{hang} synthesizes a grasp by segmenting the object into small areas, analyzed according to the curvature of the surface, that can be used for contacts with the fingertip. Using tactile feedback, the grasp configuration on the object can be changed, to increase stability, by moving one finger to a new contact point. However, this change of the contact point does not lead to a new object configuration: it is only a more stable version of the previous grasp.

In \cite{sundaralingam} the authors address the problem of planning joint motions of a multi-fingered hand to change the pose of the object inside it. Their method, called relaxed-rigidity constraint, has the advantage of relying only on the kinematics, without the need for the knowledge of dynamic properties of the object. However, it achieves only small changes in the pose of the objects. Moreover, the repositioning is not obtained by changing the grasp on the object because the contact points are kept stable. Therefore, the correction of the object pose with respect to the end-effector relies on the motion of the finger joints, which are required to have more than 1 DoF.

Similarly, in \cite{psomopoulou} the authors achieve a stable grasp by moving the two fingers with 3 DoF each to reach a new configuration of the object.

A work that instead focuses on simple parallel grippers is the one described in \cite{chavan-dafle_4}, which also addresses the problem of planning a sequence of motions for repositioning the object. The extrinsic dexterity is provided by multiple external fixtures that have been designed for proper object sliding and rotation. The sequence of pushes against these fixtures necessary to change the grasp is produced by a sampling-based planner. This planner uses a combination of RRT* and an inverse dynamic solver that, given the object's geometry, its dynamics properties, and the gripping force, models the contact between the object and the gripper or the external pusher. For a successful action, this work relies on an accurate dynamic model, including contact models, and predefined additional supports in the environment.

Strategies that allow for a simpler change in the grasp with a parallel gripper have also been explored. Pivoting \cite{sintov_2, cruciani} is the rotation of the object around a single axis. Sliding \cite{shi_2} is the planar translation of the object. By combining these two strategies, it is possible to obtain a wider range of changes in the object's configuration.

In our approach, the DMG plans a sequence of simple motions, named manipulation primitives, whose combination moves the object inside the gripper to the desired pose.

\section{Dexterous Manipulation Graph} \label{sec_dexterous_manipulation_graph}
The Dexterous Manipulation Graph $\mathrm{DMG}{=}\!<\!\!N,\: E_\mathrm{DMG}\!\!>$ is a disconnected undirected graph. A node, or vertex, $n_{ij}{=}\!<\!\!\textbf{p}_i, \: A_{ij}\!\!>\:\in N$ is a tuple, containing information about the position of the contact point $\textbf{p}_i$ and about possible finger's orientations in that contact point $A_{ij}$. An edge $e_{n_{ij}n_{hk}}\in E_\mathrm{DMG}$ connects two nodes $n_{ij}$ and $n_{hk}$ if it is possible for the fingers to move from one to the other using the manipulation primitives.

Section \ref{sec_manipulation_primitives} defines these manipulation primitives. Then, we explain how to generate the DMG in section \ref{sec_graph_generation} and analyze the complexity of the proposed algorithm in section \ref{sec_complexity_analysis}. The graph search for obtaining a sequence that moves the object to the goal pose is described in section \ref{sec_graph_search} and the obtained path of the contact points on the object is summarized in section \ref{sec_in-hand_manipulation_path}. Finally, section \ref{sec_manipulability_analysis} presents the DMG as a tool to analyze the manipulability of an object.

\subsection{Manipulation Primitives}\label{sec_manipulation_primitives}
\begin{figure}[b]
	\centering
	\begin{subfigure}[t]{0.33\columnwidth}
		\centering
		\includegraphics[width=0.65\textwidth]{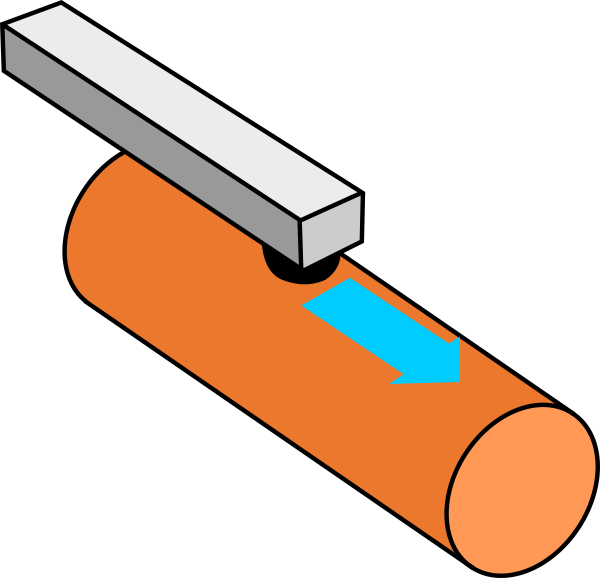}
		\caption{Translation}\label{fig_translation}
	\end{subfigure}%
	\begin{subfigure}[t]{0.33\columnwidth}
		\centering
		\includegraphics[width=0.65\textwidth]{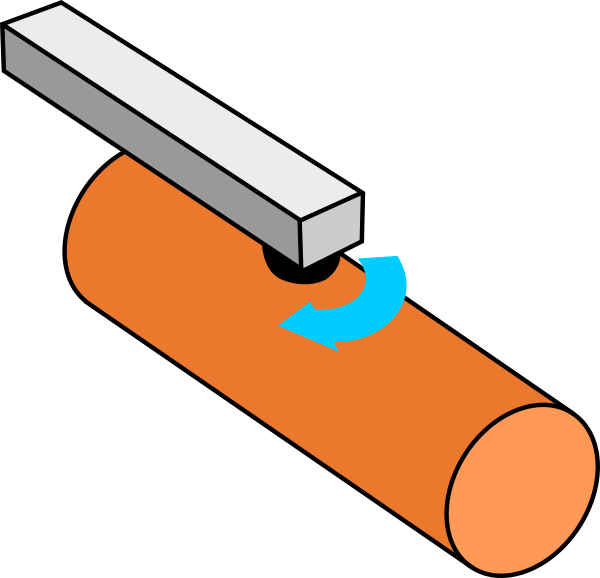}
		\caption{Rotation}\label{fig_rotation}
	\end{subfigure}%
	\begin{subfigure}[t]{0.33\columnwidth}
		\centering
		\includegraphics[width=0.65\textwidth]{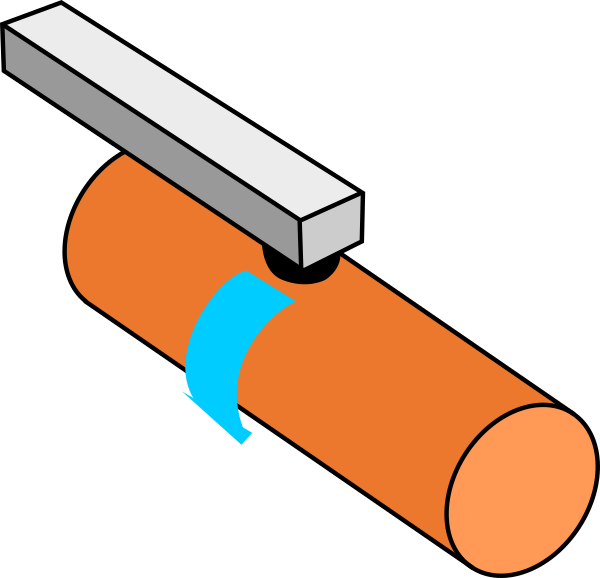}
		\caption{Rototranslation}\label{fig_rototranslation}
	\end{subfigure}
	\caption{The three manipulation primitives that describe the motion of the contact point between the fingertip and the object. The finger of a parallel gripper is represented in gray, and its fingertip in black.}
\end{figure}

In our method, the object is controlled to move inside the gripper using a combination of simple motions. These motions are defined with respect to the contact point between the fingertip and the object. Their execution requires an adjustment of the pressure exerted by the finger on the object to allow it to move, and external forces that drive the motion.

For the purpose of dexterous manipulation by changing the contact points between the fingers and the object, we have identified the following manipulation primitives:
\begin{enumerate}
	\item Translation (Fig.~\ref{fig_translation}): in this motion, the contact point slides between two points $\textbf{p}_i$ and $\textbf{p}_h$ on the surface of the object and the orientation of the object's reference frame with respect to the gripper's reference frame remains constant.
	\item Rotation (Fig.~\ref{fig_rotation}): in this motion, the contact point is fixed, but the object's reference frame changes by a planar rotation $r$ around the contact point.
	\item Roto-translation (Fig.~\ref{fig_rototranslation}): in this motion, the contact point slides between two points on the object's surface, and the orientation of its reference frame changes at the same time, not limited to planar rotation changes.
\end{enumerate}
The first two motions can be obtained with a non-prehensile pushing on the object in certain directions. A subset of motions falling in the third category can be described as a sequence of the first two. The remaining roto-translations require additional torque that is more difficult to provide. For instance, the second arm of the robot can grasp the other end of the object and move it while the first gripper adjusts the gripping force to allow the contact point to slide. 

In this work, we focus on motions that do not require additional grasp planning and multiple grasps on the object. Therefore, we designed the DMG to provide a solution that is composed only of the first two primitives. However, we aim to provide a solution for problems that involve multiple grasps and regrasps in our future developments of this work.

In our implementation, the manipulation primitive is executed by providing an additional contact point on the object using the second arm of the robot and pushing with one or both arms. The description of the dual arm motions for this execution is presented in section \ref{sec_dual_arm_execution}. 

\subsection{Graph Generation} \label{sec_graph_generation}

\begin{figure}[t]
	\centering
	\includegraphics[width=0.45\textwidth]{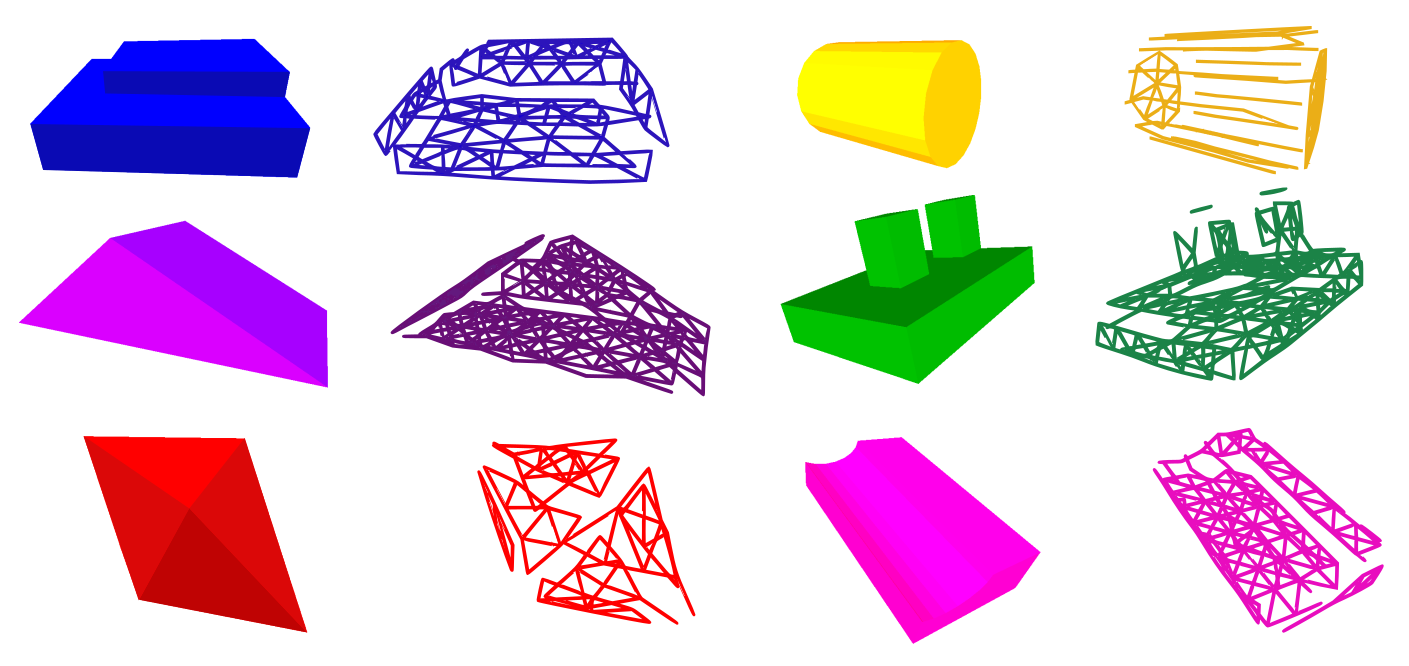}
	\caption{Examples of Dexterous Manipulation Graphs for different objects.}\label{fig_example_graphs}
\end{figure}

The DMG is generated by analyzing the shape of the object, represented as a point cloud. This point cloud is segmented into small areas, and their adjacency is analyzed and refined.
The example graphs in Fig.~\ref{fig_example_graphs} represent only a partial visualization of the information contained in the DMG. However, this visualization is intuitive and shows a rough example of what the graph provides.

The first step towards obtaining the DMG is dividing the object into small areas. To segment the object into connected areas, we use the method of supervoxels for point clouds \cite{papon}. After selecting the desired resolution, chosen according to the object's dimensions and the desired precision of the graph, this segmentation subdivides the object into areas that are grouped according to common characteristics, such as the geometric information. To each area, called supervoxel, correspond a centroid, which is a point on the surface of the object, and a normal to this surface. The centroid of the i-th supervoxel, $\textbf{p}_i$, is considered a possible contact point between the fingertip and the object. 

Each supervoxel is adjacent to its neighboring areas, and we keep this adjacency information in an initial connected graph $G=<\!V, E\!>$. Its vertexes $v\in V$ correspond to points $\textbf{p}_i$ on the object. In contrast, the nodes of the final DMG contain additional information other than the centroid of the corresponding supervoxel. The process for generating the DMG from $G$ is summarized in Algorithm \ref{algorithm} and explained below.

\begin{algorithm}[t!]
	\small
	
	\SetNoFillComment
	\DontPrintSemicolon
	\SetKwInOut{Input}{Input}
	\SetKwInOut{Output}{Output}
	
	\Input{$G$, $l$, $a_s$}
	\Output{DMG, $C_{\mathrm{DMG}}$}
	$V\leftarrow$ vertices of $G$\tcp*{$v\in V$ is a point on the object}
	$E\leftarrow$ edges of $G$\tcp*{$e_{ij}$ edge between $i$ and $j$}
	
	\ForEach{$v\in V$}{
		$\hat{\textbf{n}}_v\leftarrow$ normal in $v$ \\
		\ForEach{$w\mathrm{~adjacent~to~} v$}{
			$\hat{\textbf{n}}_w\leftarrow$ normal in $w$ \\
			\If{$||\hat{\textbf{n}}_v-\hat{\textbf{n}}_w||>\delta$}{
				$E\leftarrow E\smallsetminus\{e_{vw}\}$ 	
			}
		}
		\If{$v \mathrm{~is~isolated}$}{
			$V\leftarrow V\smallsetminus\{v\}$\\	
		}
	}
	$C_G\leftarrow$ connected subgraphs in $G$\\
	$N\leftarrow\varnothing$ \tcp*{empty set of nodes}
	\ForEach{$v \in V$}{
		$A\leftarrow\varnothing$ \tcp*{empty set of admissible angles}
		\ForEach{$a \in \{0: a_s: 360\}$}{
			\If{$a \mathrm{~is~admissible~with~finger~length~} l$}{
				$A\leftarrow A\cup \{a\}$ \\
			}
		}
		\eIf{$A \mathrm{~is~empty}$}{
				remove $v$ from $V$ and from its component $C_v\in C_{G}$
			}
		{
			$A_v\leftarrow\varnothing$ \tcp*{empty set of angle components}
			$j\leftarrow 0$\\
			\ForEach{$\mathrm{sequencial~set~in~} A$}{
				$A_{vj}\leftarrow$ j\textsuperscript{th} sequencial set in $A$\\
				$n_{vj}\leftarrow <v, A_{vj}>$\\
				$N\leftarrow N\cup\{n_{vj}\}$\\
				$A_v\leftarrow A_v\cup \{A_{vj}\}$\\
				$j$++\\
			}
		}	
	}
	$E_{\mathrm{DMG}}\leftarrow$ empty set of edges\\
	\ForEach{$C_i\in C$}{
		$V_i\leftarrow$ points in $C_i$\\
		$E_i\leftarrow$ connections in $C_i$\\
		\ForEach{$v \in V_i$}{
			\ForEach{$A_{vj}\in A_v$}{
				\ForEach{$w\mathrm{~adjacent~to~} v \mathrm{~in~} C_i$}{
					\ForEach{$A_{wk}\in A_w$}{
						\If{$A_{vj}\cap A_{wk}\neq\varnothing$}{
							$E_{\mathrm{DMG}}\leftarrow E_{\mathrm{DMG}}\cup \{e_{n_{vj}n_{wk}}\}$
						}	
					}
				}
					
			}	
		}
	}
	DMG$\leftarrow<N, E_{\mathrm{DMG}}>$\\
	$C_{\mathrm{DMG}}\leftarrow$ connected subgraphs in DMG	 	
	
\Return DMG, $C_{\mathrm{DMG}}$

\caption{graph\_generation}\label{algorithm}

\end{algorithm}

Once the object has been subdivided, a first refinement of the connections is performed. Given the normals $\hat{\textbf{n}}_i$ and $\hat{\textbf{n}}_h$ of the i-th and h-th adjacent supervoxels, the connection between $\textbf{p}_i$ and $\textbf{p}_h$ is removed if $\parallel\!\! \hat{\textbf{n}}_i-\hat{\textbf{n}}_h \!\!\parallel_2 >\!\delta$.
The threshold $\delta$ ensures that the fingertip does not move across sharp edges, and it can be adjusted to allow for motions on uneven or curved surfaces to the desired degree. 

After refining the connectivity, the graph is divided into separate connected components $C_G$, on which the fingertip can slide and rotate while maintaining contact. However, we need to take into account the finger's body, and not just the fingertip, when  executing different movements on the objects. Therefore, we further refine the adjacency assuming a finger of length $l$ of a parallel gripper. In fact, collisions can occur due to non convex shapes and due to objects being longer than the finger.

We discretize the possible orientations that a finger can assume in the set $[0\degree, 360\degree)$ with a step $a_s$. A finger can be in contact with the object at a point $\textbf{p}_i$ and it can assume only orientations that are not in collision. These orientations belong to the set of possible angles $A_i$. If this set is empty, the point is removed from the graph because it is impossible for the gripper to grasp there.

The set of possible angles $A_i$ is then further analyzed to finally obtain the DMG and the connection between its nodes. The angular component $A_{ij}\subseteq A_i$ of a node $n_{ij}$ is a set of admissible orientations that are in a continuous sequence among the ones in $A_i$. More specifically, the finger is able to rotate from the minimum angle in $A_{ij}$ to the maximum angle using a single continuous rotation, without encountering collisions. A connection between two nodes $n_{ij}$  and $n_{hk}$, corresponding to the contacts $\textbf{p}_i$ and $\textbf{p}_h$, is inserted in the DMG only if $A_{ij}\cap A_{hk}\neq\varnothing$.

\begin{figure}[t]
	\centering
	\includegraphics[width=0.25\textwidth]{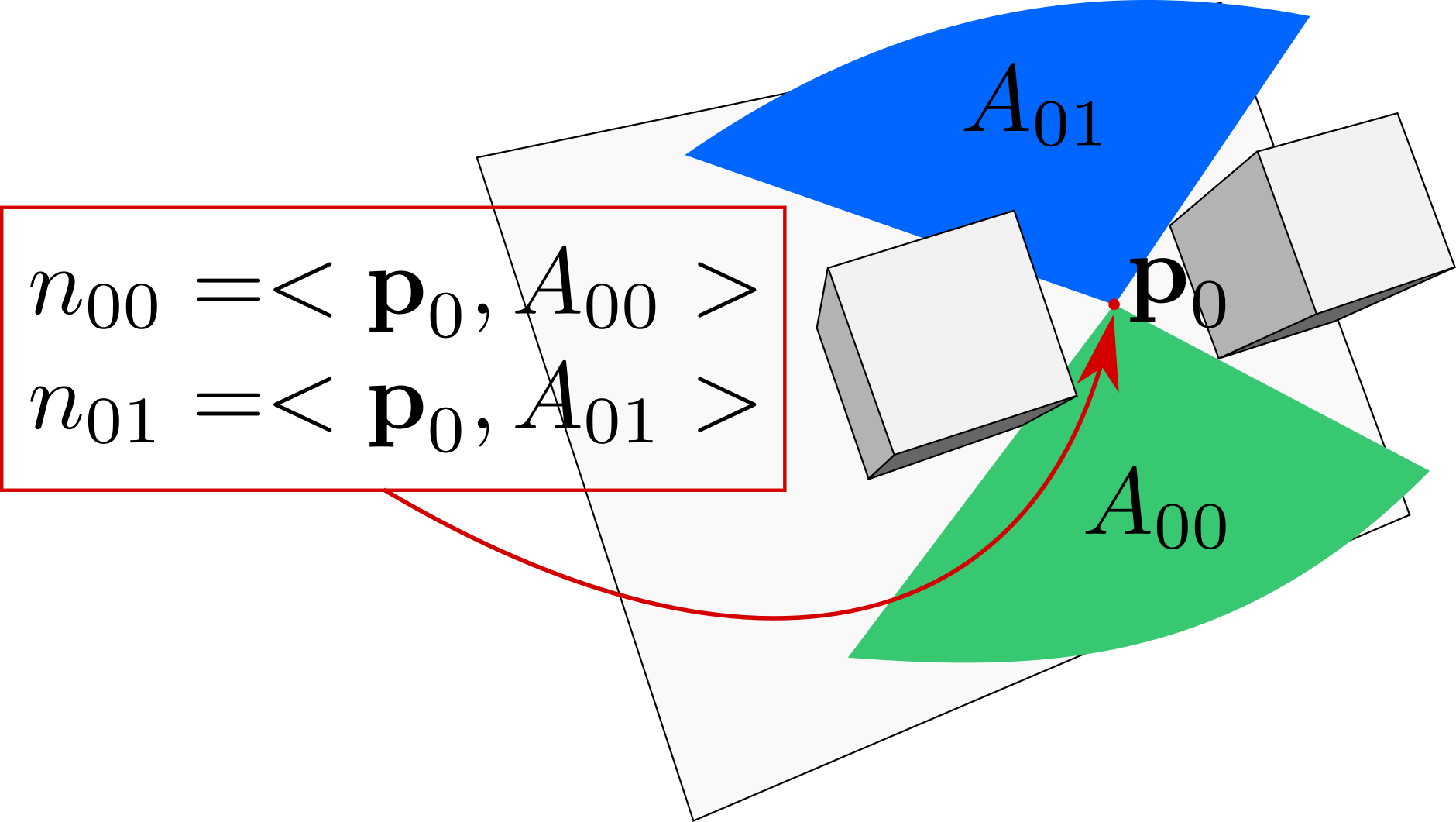}
	\caption{The point $\textbf{p}_0$ on the surface of this plug-shaped object corresponds to two different and disconnected nodes in the DMG, $n_{00}$ and $n_{01}$, because a gripper's finger can grasp it with two separate angle ranges, $A_{00}$ and $A_{01}$.}\label{fig_angular_components}
\end{figure}

If an area has multiple angular components, the corresponding nodes in the DMG are disconnected because the finger cannot rotate at that contact point between the two disjoint sets of orientations. In fact, a contact occurring in the position $\textbf{p}_i$ on the object's surface can correspond to more than one node in the graph, because there is a disconnection in the angular components, as in the example in Fig.~\ref{fig_angular_components}.

When refining the adjacency by taking into account the orientation, the amount of connected components in the graph can increase with respect to the ones in $C_G$, depending on the object's geometry. Therefore, we define the set of connected components in the DMG as $C_{\mathrm{DMG}}$.

\subsection{Complexity Analysis} \label{sec_complexity_analysis}

The complexity of the generation of the DMG can be easily derived. Although the process in Algorithm \ref{algorithm} can be optimized, we prefer this notation for readability purposes, and we assume a Breadth-first search for the operation of deriving the connected components. 

Given $|V|$ as the number of vertexes in $G$, which depends on the chosen resolution in the segmentation of the object, $a$ as the number of considered angles, which is $\left\lceil\frac{360}{a_s}\right\rceil$, $b$ as the maximum number of edges from a node, and $c$ as the maximum number of angular components, the complexity is

\begin{equation}
	O((\max(c^2 b,\:ac))|V|),
\end{equation}
where $b$ is usually at most 4 or 5 because of the geometric adjacency of the supervoxels, $c$ is 1 for most of the nodes and it can be expected to be a small number (2 or 3) for a few of them, only for unusual and complex shapes, and $a$ depends on the chosen resolution for the angles ($a_s$).

\subsection{Graph Search}\label{sec_graph_search}

The DMG is composed of separate connected components, in which the contact point can move. If the desired contact on the object lies in the component $C_D \in C_{\mathrm{DMG}}$, and this component is different from the component $C_I \in C_{\mathrm{DMG}}$  of the initial contact, it is not possible to achieve the desired repositioning without releasing the contact between the gripper and the object. 

\begin{figure}[b]
	\centering
	\includegraphics[width=0.3\textwidth]{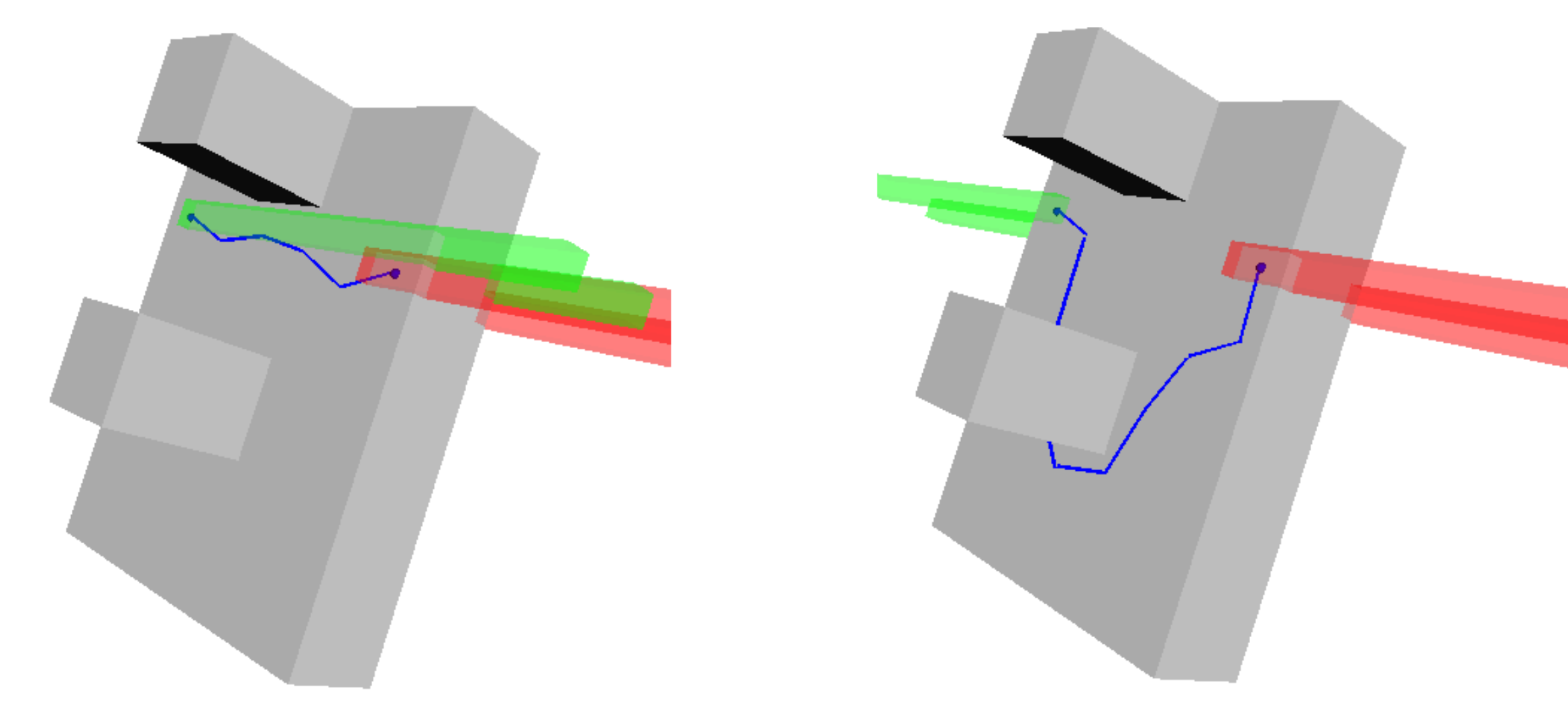}
	\caption{A path for the contact point to move the gripper from the initial configuration (in red), to the goal configuration (in green). The contact points are the same, but the two goals have a different gripper's orientation. The different angular components provide two different paths.}\label{fig_path_example}
\end{figure}

When searching for a path within a connected component, optimal and fast strategies for graph search can be used. In our particular implementation, we use Dijkstra's algorithm. Fig.~\ref{fig_path_example} shows an example of two paths. These paths are found between the same contact points, but with two different target orientations of the gripper.

We address the problem of in-hand manipulation with a parallel gripper, which has two contacts on the object. However, the DMG is built taking into account only one contact point. The contact points of the two fingers on the object are associated to the node with the closest position and with angular components that contains the current orientation. 
We assume that one finger is the \emph{principal finger}, arbitrarily chosen, and the other one, the \emph{secondary finger}, follows it on the opposite side in its own corresponding connected component of the DMG. When looking for a feasible solution, both fingers must remain in their component during the whole motion, named $C_p$ for the principal finger and $C_s$ for the secondary finger. 

The graph search algorithm is executed from the initial principal finger contact $n^p_{init}$ to the desired one $n^p_{goal}$. At each iteration, to a node $n^p_{ij}$ in $C_p$ corresponds a secondary finger's contact point. This point is found as the intersection between the object and the line starting from the position of the principal contact $\textbf{p}_i$ with the same direction of the line that connects the two fingers in the previous configuration. In fact, a motion between nodes with different positions of the contact point corresponds to the translation movement primitive, and therefore the orientation of the fingers with respect to the object remains constant. The obtained point is associated to a node $n^s_{xy}$ in the DMG. If $n^s_{xy}$ is not in $C_s$, or the gripper is in an invalid configuration, the distance between the two considered principal nodes $n^p_{ij}$ and $n^p_{hk}$ is set to $\infty$ to ensure that this path is not chosen.

\begin{figure}[t]
	\centering
	\includegraphics[width=0.4\textwidth]{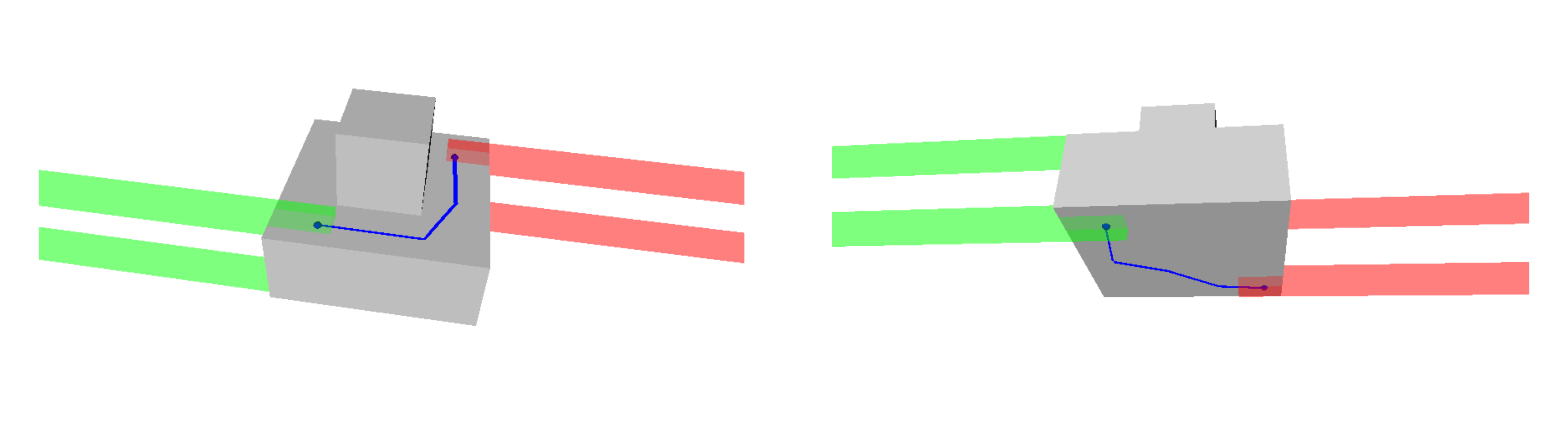}
	\caption{An example of the same in-hand repositioning computed assigning a different gripper's finger as principal. The found paths are both valid, but they lead to different motions of the object.}\label{fig_example_master_slave}
\end{figure}

The obtained path may not be the shortest path for the principal finger, because it depends also on the secondary finger on the opposite side of the object. As shown in Fig.~\ref{fig_example_master_slave}, in the second image the path avoids the center of the object, although for that contact point alone moving there would be the optimal solution.

As final comment, we briefly discuss how to use the distance between two nodes $n^p_{ij}$ and $n^p_{hk}$ during the search. This distance can be the simple euclidean distance $\parallel\! \textbf{p}_h - \textbf{p}_i \!\parallel_2$, or it can be chosen according to how expensive a movement is with the available robot. For instance, if a rotation is required because the current angle $\gamma$ of the finger is so that $\gamma \in A_{ij}$ but $\gamma \notin A_{hk}$, this rotational movement can be considered as more expensive by increasing the distance.

In addition, the parallel gripper can increase or decrease the opening between the two fingers when it is manipulating an object whose surfaces are not parallel. This opening is obtained from the distance between the contacts of the principal and secondary fingers. Depending on the precision and efficiency in actuation of the available gripper, changes in distance between the fingers can be penalized during the graph search, according to how preferable it is to keep a stable fingers' opening versus a possibly longer path.

Finally, some motion directions can be also penalized by increasing the distance between the corresponding nodes, to avoid undesirable behaviors in the execution, such as the examples mentioned later in sections \ref{sec_translation_execution} and \ref{sec_rotation_execution}.

\subsection{In-hand Manipulation Sequence} \label{sec_in-hand_manipulation_path}

The path obtained from the DMG is a sequence of $K$ nodes $\{n^0, n^1, ..., n^{K-1} \}$, with $n^0{=}n^p_{init}$ and $n^{K-1}{=}n^p_{goal}$. This path must be translated into a sequence of manipulation primitives. 

Given the nodes $n^k{=}\!\!<\!\!\textbf{p}^k, A^k\!\!\!>$ and $n^{k+1}{=}\!\!<\!\!\textbf{p}^{k+1}, A^{k+1}\!\!\!>$ in sequence in the path and the current orientation of the finger $\gamma^k$ in the contact point of the node $n^k$, two manipulation primitives are needed to move the contact from $n^k$ to $n^{k+1}$:
\begin{enumerate}
	\item A rotation $r^k$ so that the next orientation $\gamma^{k+1}{=}\gamma^{k}+r^k$ is so that $\gamma^{k+1}\in A^k\cap A^{k+1}$. This rotation can be chosen to minimize the change, so that it is $0$ when $\gamma^{k}\in A^k\cap A^{k+1}$, to get closer to the goal angle $\gamma^{K-1}$, or to fulfill additional requirements. 
	\item A translation $\textbf{t}^k$ that slides the contact point on the object from $\textbf{p}^k$ to $\textbf{p}^{k+1}$. This translation is obtained as $\textbf{t}^k{=}\textbf{p}^{k+1}-\textbf{p}^k$. Therefore, from a path of $K$ nodes, the corresponding translations are $K-1$.
\end{enumerate}

The sequence of manipulation primitives is an alternation of rotations and translations, plus a final rotation to reach the final angle: $\{r^0, \textbf{t}^0, r^1, \textbf{t}^1, ..., r^{K-2}, \textbf{t}^{K-2}, r^{K-1}\}$. It is possible to have several rotations with value $0$, which can then be ignored during the execution. In contrasts, the translations are always nonzero because there is no connection between nodes $n_{ij}$ and $n_{ik}$, with the same contact point but different angular component. Since the path is fragmented into many segments, we unite these segments to avoid a slow execution. For instance, two translations are joint if there is a rotation equal to $0$ in between and the change in direction of the path on the object's surface is negligible.

\subsection{From DMG To Manipulability Analysis} \label{sec_manipulability_analysis}
In addition to planning dexterous manipulation, the DMG can be used to analyze the manipulability of an object. In particular, we focus on how much the manipulation primitives can reposition an object inside the gripper without the need for regrasping.

We build a \textit{manipulability matrix} by computing a path between several different poses of the gripper, distributed in the object's bounding box. This matrix contains $1$ in the $ij$ position if a path exists between pose $i$ and pose $j$, and $0$ otherwise. The resulting matrix is symmetric and it is an adjacency matrix. By setting the diagonal elements to $0$, loops between nodes are removed. This matrix can be seen as an extended representation of the DMG, but it strictly depends on the chosen parallel gripper, while in the DMG each node is associated to a single contact point and this abstraction is an advantage in case our method is extended to multi-fingered hands. 

This matrix can be ordered into blocks, which are a simple tool to visualize the manipulability of the objects. The manipulability matrix provides an overview on how easy it is for the robot to manipulate a certain object without the need to regrasp. In fact, if there are fewer blocks, the likelihood of manipulating an object while maintaining contact is higher. This matrix can also be used in future implementations as an aid to a regrasping strategy that takes into account both the object's shape and the current gripper's configuration, because it provides information on an area where to regrasp. 
Examples of manipulability matrices are analyzed in section \ref{sec_experiments_objects_manipulability}.

\section{Dual Arm Formulation} \label{sec_dual_arm_execution}
As mentioned, we use the ECTS \cite{park_1} to describe the motion of our dual arm system. The ECTS allows for a specification of the absolute and relative velocities of the two end-effectors and it translates them into velocities for each robot's arm. Additionally, the degree of coordination of each arm can be adjusted to obtain symmetrical or asymmetrical executions.

With $\dot{\textbf{x}}{=}(\dot{\textbf{x}}_1, \: \dot{\textbf{x}}_2)^T$ being the vector containing the Cartesian velocities of the first and second end-effectors, and $\dot{\textbf{x}}_a{=}(\textbf{v}_a, \: \bm{\omega}_a)^T$ and $\dot{\textbf{x}}_r{=}(\textbf{v}_r, \: \bm{\omega}_r)^T$ being their absolute and relative motions expressed as linear and angular velocity, the relationship between these velocities is
\begin{equation} \label{eq_ects}
	\dot{\textbf{x}}=\left[\begin{array}{cc}
	\alpha\textbf{I}_6 & -(1-\alpha)\textbf{I}_6 \\
	-\beta\textbf{I}_6 & \alpha\textbf{I}_6
	\end{array}\right]\dot{\textbf{x}}_E,
\end{equation}
where $\dot{\textbf{x}}_E{=}(\dot{\textbf{x}}_a, \: \dot{\textbf{x}}_r)^T$, $\textbf{I}_6$ is the 6-dimensional identity matrix, and $\alpha$ and $\beta$ are coordination coefficients. In particular, $\beta\in\{0, 1\}$ and $\alpha\in[0, 1]$. With $\beta$ set to $0$, the motion is uncoordinated, therefore in our case it is always set to $1$, as the two arms are manipulating the same object with the purpose of repositioning it inside one of the two grippers. $\alpha$ defines the degree of coordination. For instance, if it is set to $0.5$, the motion of the arms is symmetrical. If it is set to $0$, only one arm moves and this mode can be seen equivalent to a system of a single arm robot and an external support.

The motion of the object inside the gripper can be described using the relative motion of the two arms $\dot{\textbf{x}}_r$. In our experiments, we keep the absolute motion $\dot{\textbf{x}}_a$ to zero, but it can be exploited to move the arms while the robot is manipulating the object.

In this section, we first provide an overview of the execution procedure and then we explain in detail how to obtain the translation and rotation manipulation primitives with the dual arm system.

\subsection{Execution Procedure}

In our dual arm system, the robot grasps the object with its first gripper, and its second gripper helps pushing the object to adjust its pose. The object must be moved according to the sequence of manipulation primitives derived in section \ref{sec_in-hand_manipulation_path}. To each translation $\textbf{t}^k$ and to each rotation $r^k$ correspond the following steps executed by the robot:
	\subsubsection{Find contact point} 
\begin{figure}[b]
		\centering
		\begin{subfigure}[t]{0.49\columnwidth}
			\centering
			\includegraphics[width=0.9\textwidth]{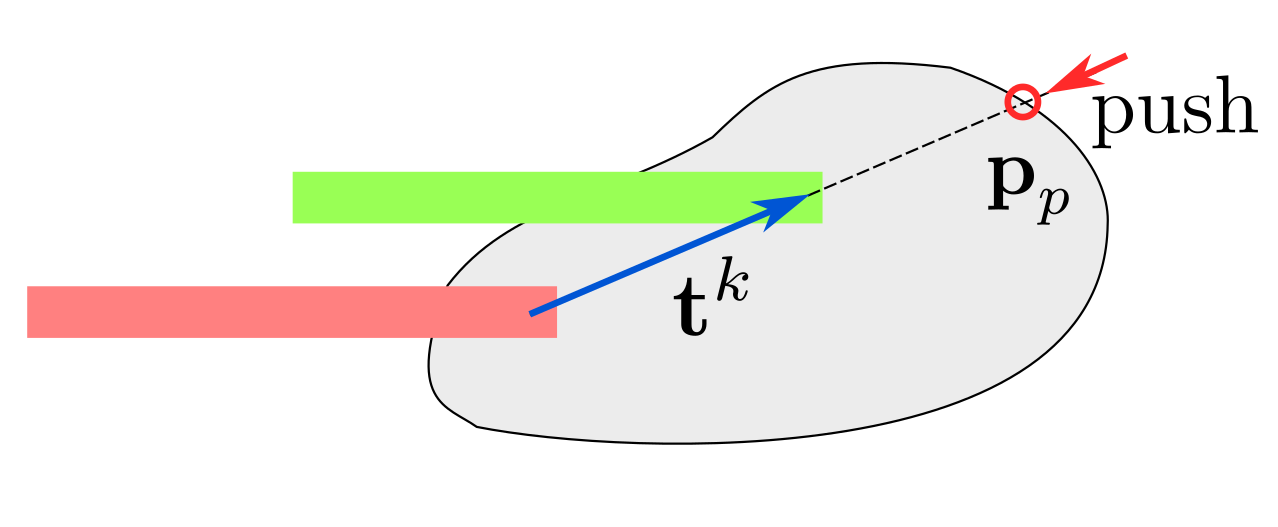}
			\caption{Given $\textbf{t}^k$, the object is pushed from a point $\textbf{p}_p$ found as the intersection between its surface and the line passing through the contact point with the direction of the translation.}\label{fig_translation_contact}
		\end{subfigure} 
		\begin{subfigure}[t]{0.49\columnwidth}
			\centering
			\captionsetup{width=.95\linewidth}
			\includegraphics[width=0.9\textwidth]{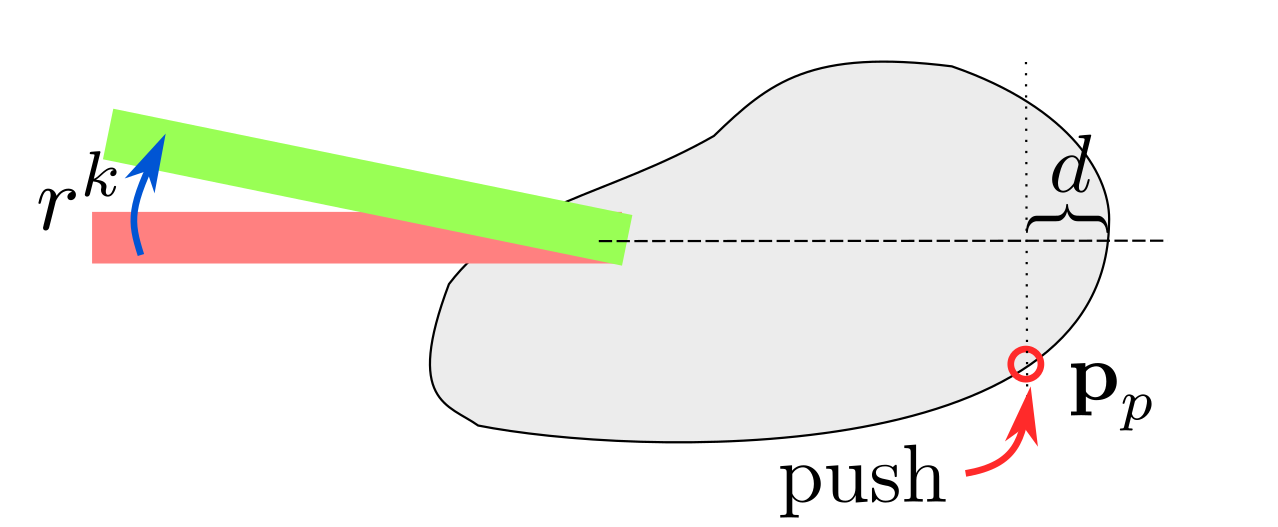}
			\caption{Given $r^k$, $\textbf{p}_p$ is one of the two intersections between a line orthogonal to the finger, put inside the object of a distance $d$, chosen according to the direction of $r^k$.}\label{fig_rotation_contact}
		\end{subfigure}
		\caption{The derivation of the contact point between the second gripper and the object. The initial finger's pose is red, and the desired one is green.}\label{fig_contact_point}
	\end{figure}

In this step, the robot does not move, but it analyzes the shape of the object. In fact, for the second gripper to be able to push the object in the desired pose, the pushing must begin from a proper contact point with the object. Fig.~\ref{fig_contact_point} shows examples on how to find the contact point between the object and the second gripper.
	\subsubsection{Approach contact point} In this step, the robot moves its arms so that the second gripper is in contact with the object at the chosen point.
	\subsubsection{Execute manipulation primitive} In this step, the robot moves the arms so that the object's pose inside the first gripper changes according to the corresponding manipulation primitive. The first gripper also enlarges and tightens the fingers to facilitate the object's motion. The explanation of the translation execution is given in section \ref{sec_translation_execution}, and the rotation execution is described in section \ref{sec_rotation_execution}.
	\subsubsection{Leave contact point} In this step, the robot moves the arms so that the second gripper is no longer in contact with the object. The first gripper is instead still grasping the object, in the new pose that derived from the execution of the manipulation primitive.

\subsection{Translation Execution} \label{sec_translation_execution}

If the manipulation primitive to be executed is translation, then
\begin{equation} \label{eq_translation_velocity}
	\dot{\textbf{x}}_r=\left(\textbf{v}, \:\textbf{0}\right)^T.
\end{equation}
$\textbf{0}$ is a 3-dimensional zero vector, indicating that there is no angular velocity because the orientation of the object inside the gripper must remain constant. The linear velocity $\textbf{v}$ depends on the desired translational motion of the object. 

Given the translation $\textbf{t}^k$ of the contact point on the surface of the object, the direction of the translation of the contact point in the robot's base frame is $\textbf{t}'^k{=}\;^b\textbf{R}_{g_1}\tilde{\textbf{t}}^k$, where $^b\textbf{R}_{g_1}$ is the rotation matrix containing the orientation of the gripper in the base reference frame and $\tilde{\textbf{t}}^k$ is a 3-dimensional vector containing the 2-dimensional $\textbf{t}^k$ expressed in the gripper's reference frame. To achieve a translation of the contact point directed as $\textbf{t}'^k$, the object must move in the opposite direction, denoted as ${-}\hat{\textbf{t}}'^k$. We adjust the magnitude of the velocity $m$ using a P controller during the motion execution, until the fingertip reaches the desired contact point. In conclusion, the linear velocity is $\textbf{v}{=}{-}m\hat{\textbf{t}}'^k$.

We assume that the friction between the object and the second gripper is so that the latter can push the object also along diagonal directions. However, since the DMG is independent from the initial grasping configuration, it contains translational motions along directions that may not be feasible for pushing, and which instead require pulling. A pulling motion would require the second gripper to grasp the object, and since we do not deal with regrasping problems in this work, we prefer to discard translational motions that require a pulling motion at the planning stage. More specifically, these motions are penalized in the cost of the path during the graph search, so that a preferred solution contains a rotation of the object, a push and a second rotation rather than a pull.

\setcounter{figure}{9}
\begin{figure*}[b]
	\centering
	\includegraphics[width=\linewidth]{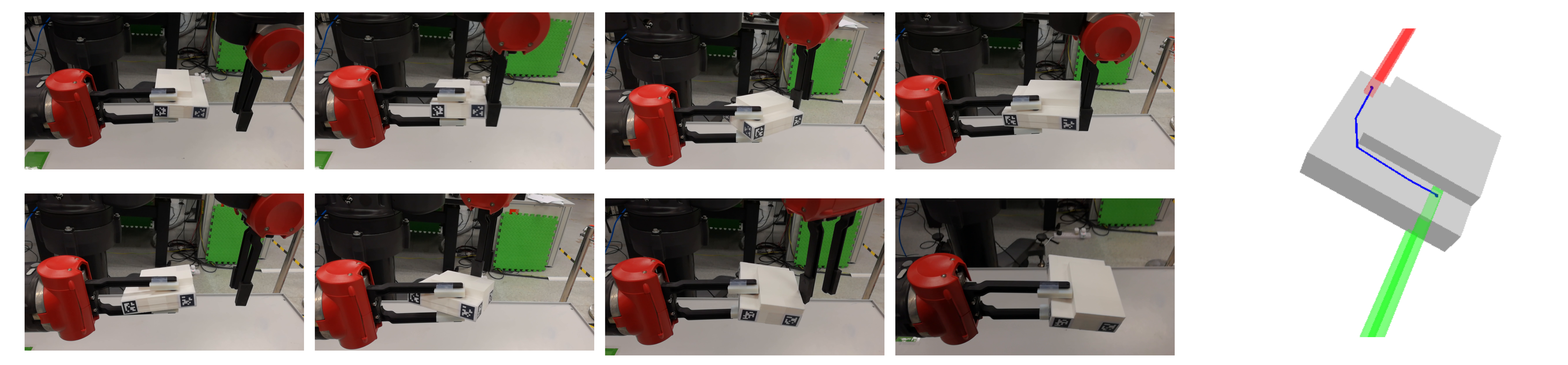}
	\caption{The in-hand manipulation executed with Baxter. The motion of the contact point is shown on the right. The execution is composed of an alternation of four rotations (of which the first two are $0$) and three translations: the second gripper approaches the object and pushes it for translating the contact points with the first gripper, then it changes the approach point and the push direction to impose a rotation on the object, and it repeats these motions until the object reaches the desired pose.} \label{fig_execution_baxter}
\end{figure*}

\subsection{Rotation Execution} \label{sec_rotation_execution}

If the motion to be executed is a rotation, the relative motion of the second gripper's contact point is along an arc of a circle. We induce also an angular velocity so that the second gripper follows the object during the rotation without altering the contact surface.

\setcounter{figure}{7}
\begin{figure}[t]
	\begin{minipage}[b]{.15\textwidth}
		\centering
		\captionsetup{width=.95\linewidth}
		\includegraphics[width=\textwidth]{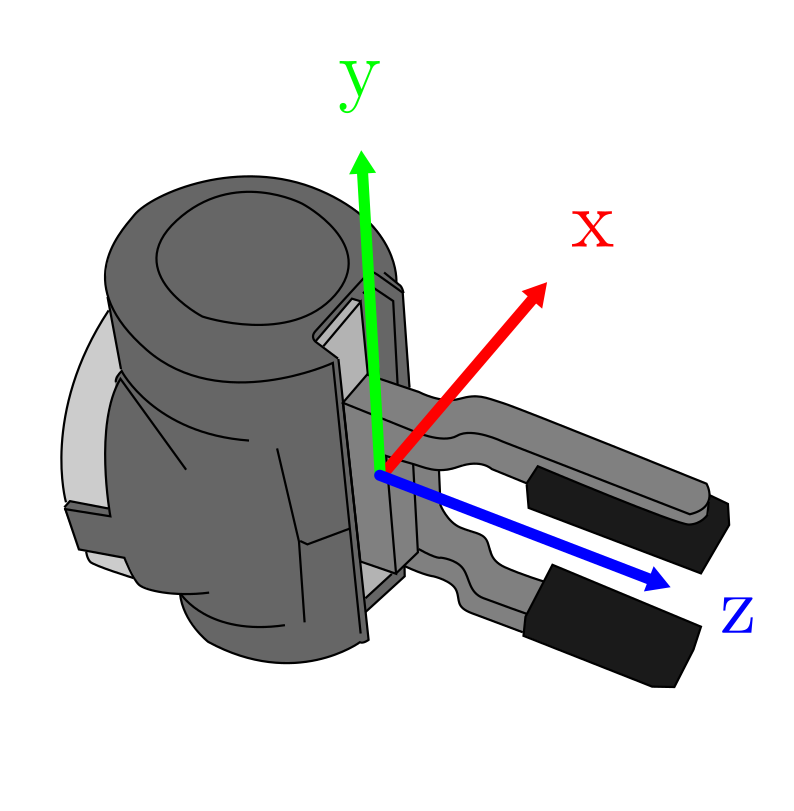}
		\caption{The gripper's reference frame.}\label{fig_gripper_frame}
	\end{minipage}%
	\begin{minipage}[b]{.35\textwidth}
		\centering
		\captionsetup{width=.95\linewidth}
		\includegraphics[width=0.6\textwidth]{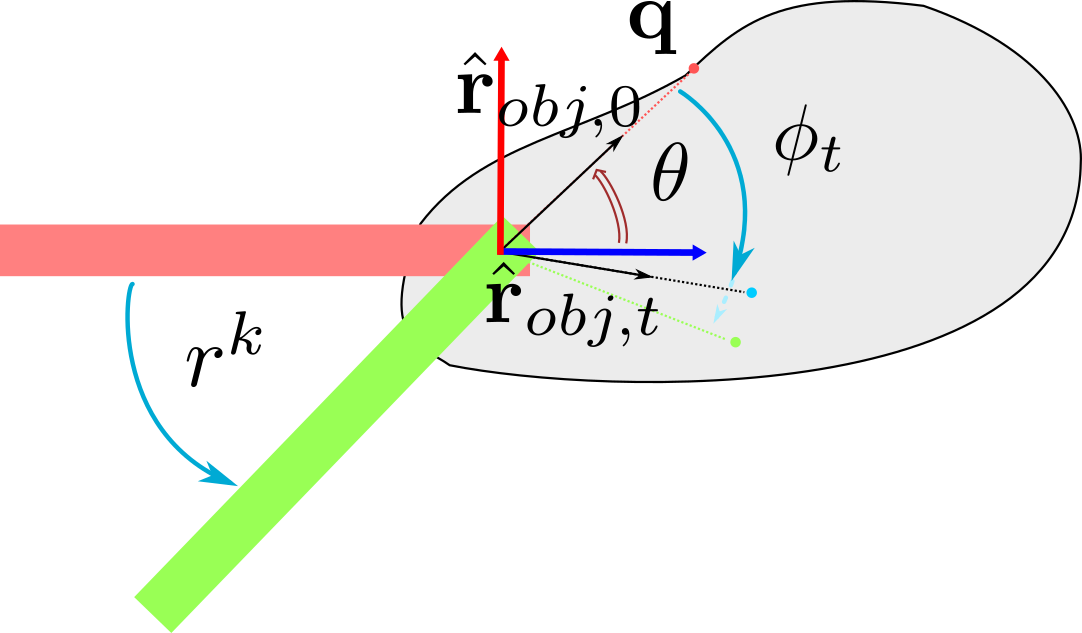}
		\caption{A rotation $r^k$, from a finger's initial orientation (in red) to the desired one (in green).}\label{fig_object_rotation}
	\end{minipage}
\end{figure}

Assuming the reference frame of the gripper as in Fig.\ref{fig_gripper_frame}, the velocity in the first gripper frame is
\begin{equation} \label{eq_rotation_velocity}
^{g_1}\!\dot{\textbf{x}}_r=(\cos(\theta + \phi_t)\dot{\phi}_t,\:
0,\:
-\sin(\theta+\phi_t)\dot{\phi}_t,\:
0,\:
\dot{\phi}_t, \:
0)^T.
\end{equation}
$\theta$ is the initial angle of the second gripper's contact point $\textbf{q}$ on the object with respect to the first gripper's orientation, and $\phi_t$ is its variation. Given the desired rotation of the finger in contact $r^k$, $\phi_t$ produces a rotation in the opposite direction. As represented in Fig.~\ref{fig_object_rotation}, the gripper's finger has to rotate of $r^k$, and therefore the point $\textbf{q}$ on the object rotates around the grasp contact point in the opposite direction. $\phi_t$ evolves during time so that at the beginning it is equal to $0$ and at the end it is equal to ${-}r^k$. The point $\textbf{q}$ is initially at an angle $\theta$ with respect to the gripper. $\hat{\textbf{r}}_{obj,t}$ is the unitary vector pointing in the direction of $\textbf{q}$ during the motion. Its evolution in the zx plane is described as $\hat{\textbf{r}}_{obj,t}{=}(\cos(\theta{+}\phi_t), \: \sin(\theta{+}\phi_t))^T$, from which the velocity in (\ref{eq_rotation_velocity}) is derived. This velocity is then transformed into the relative velocity $\dot{\textbf{x}}_r$ expressed in the global reference frame.

Similarly to the translation velocity, we adjust the magnitude of this rotation velocity during the execution, until the object reaches the desired orientation. Excessive rotations that could lead to the second gripper colliding with the first arm should be penalized as the pull motions for the translations.

\section{Experiments} \label{sec_experiments}
We tested our dexterous manipulation strategy on a Baxter robot. In addition, we used the DMG to evaluate the manipulability of objects, and to analyze possible aids to a regrasping strategy that compensates for desired configurations that are not reachable by moving the contact point along the object's surface.

\subsection{Robot Experiments}

We used a Baxter robot to execute the dual arm strategy defined in section \ref{sec_dual_arm_execution}. We detected the pose of the object with respect to the gripper using April tags and the camera built on Baxter's end-effector, with a frequency of $28$ fps. We used hemispherical fingertips for better controlling the friction at the contact point when enlarging or tightening the gripper's fingers for adjusting the motion of the object. 

As mentioned, the strategy consists of a sequence of pushes on the object and an adaptation of the distance between the fingers to allow the object to move. We have kept the absolute motion $\dot{\textbf{x}}_a$ to zero, leaving only the relative motion $\dot{\textbf{x}}_r$ to be executed by the two arms. 

Due to the compliant nature of Baxter, the executions of the pushing was subject to position errors. However, the accuracy was good enough to achieve the repositioning within a range of about $10$mm and $5\degree$ around the desired configuration of the gripper's finger on the object.

We have also verified that the error was smaller when the gripper that was grasping the object was kept fixed and only the second gripper was moving ($\alpha{=}1$). In the opposite case ($\alpha{=}0$), in which the first gripper moves against the second one, or in blended mode ($0{<}\alpha{<}1$), we noticed that the second gripper was significantly pushed away due to the compliance in the arm. Therefore, precise repositionings were more difficult to achieve in this dual arm mode. However, we highlight that the mode $\alpha{=}0$ is a mode that can be used to substitute the dual arm system with a single robot arm that pushes against the environment, non-compliant, instead of against its second gripper.

Fig.~\ref{fig_execution_baxter} shows an example of the execution of an in-hand repositioning. The desired final configuration is shown on the right side of the image, as well as the executed path of the contact point on the object's surface. The DMG was built using a resolution of $13$mm of the supervoxels, which produced $117$ segmented areas, and refined with $\delta{=}0.07$, $a_s{=}5\degree$ and $l{=}100$mm. The P values were $0.7$ for the fingers' distance, $0.32$ for the linear velocity and $16$ for the angular velocity.
The execution of this experiment and other experiments with different objects are available in the supplementary video.  

\subsection{Object's Manipulability} \label{sec_experiments_objects_manipulability}

We have used the DMG to analyze the manipulability of an object by building its manipulability matrix. After removing the rows and column corresponding to invalid poses, and hence containing non useful data, the matrices were ordered into blocks. Fig.~\ref{fig_matrices} shows an example of the manipulability matrix with two different objects. The correspondence of the blocks is not with a connected component of the DMG, but with an area of the object in which the gripper can manipulate it without losing contact. For instance, the bottom part of the second object is associated with both the red and the blue blocks, while it corresponds to a single connected component in the DMG. In fact, while in the DMG only one contact point is considered, the matrix contains information related to both of the gripper's fingers.
\setcounter{figure}{10}
\begin{figure}[t]
	\centering
	\includegraphics[width=0.48\textwidth]{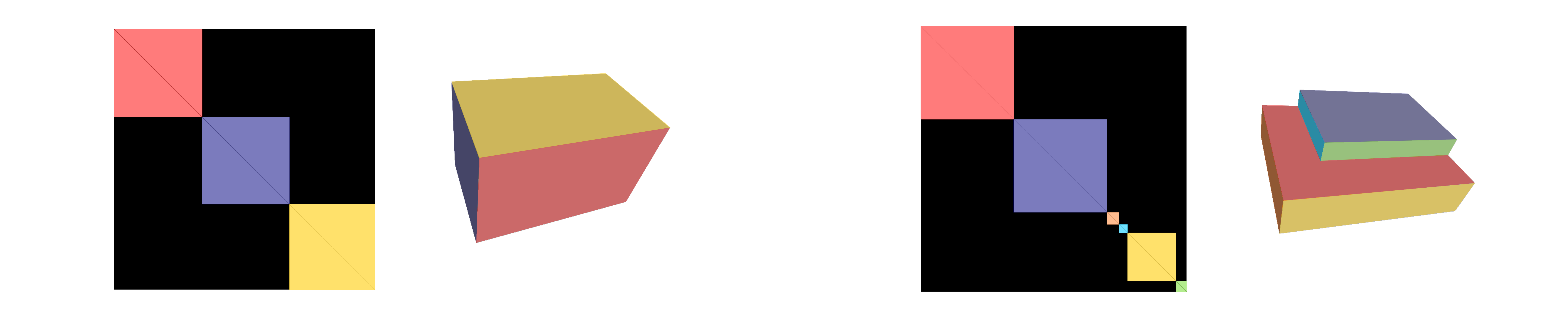}
	\caption{The manipulability matrices associated with the object's surface. The first matrix is $1332 \times1332$ and the second one is $1284\times 1284$. The matrices are binary, and black represents 0. The colors are superimposed to the blocks of 1 to show the different areas of gripper's contact.}\label{fig_matrices}
\end{figure}

The manipulability matrices also provide useful information for future implementations of a regrasping strategy, because they connect the desired pose to two connected components, but they also provide an area inside the connected component that is preferred, in case a connected component is associated to multiple blocks, hence restricting the search area for a grasp.

\section{Conclusions} \label{sec_conclusions}
We have introduced a novel tool for planning in-hand manipulation that we named Dexterous Manipulation Graph. 
First, we defined manipulation primitives that can be used by the robot to move an object inside the grasp while maintaining contact. 
Then, We suggested to analyze the shape of the manipulated object to generate a graph that describes the possible motions of the contact point between the fingertip and the object. This graph is used to search for sequences of manipulation primitives to reposition the object from an initial configuration inside the gripper to a desired one. The DMG can also be used to analyze the manipulability of an object using its manipulability matrix.
In addition, we formulated the execution of the translation and rotation movement primitives in a dual arm framework. However, the same movement can be executed with a single arm and an external contact without modifying our definitions.
Finally, we have tested our approach to in-hand manipulation using a Baxter robot.

As future work, we plan to include the possibility to grasp the object with the second arm of the robot and execute manipulations that include the roto-translation primitive. This inclusion will likely change the shape of the graph for objects without sharp edges. 
Additionally, we will include manipulations that require a regrasp, and we will use the DMG and the manipulability matrix to plan where the regrasp should be and where the second gripper should grasp so to not hinder the repositioning inside the first gripper.
Finally, we also plan to extend the DMG to be used with multi-fingered hands, and exploit the knowledge of contact-based connected components in the graph and the additional intrinsic dexterity to achieve even more dexterous in-hand manipulations.





\bibliographystyle{IEEEtran}
\bibliography{Pivoting,DexterousManip,General,DualArm}

\begin{thebibliography}{10}
\providecommand{\url}[1]{#1}
\csname url@rmstyle\endcsname
\providecommand{\newblock}{\relax}
\providecommand{\bibinfo}[2]{#2}
\providecommand\BIBentrySTDinterwordspacing{\spaceskip=0pt\relax}
\providecommand\BIBentryALTinterwordstretchfactor{4}
\providecommand\BIBentryALTinterwordspacing{\spaceskip=\fontdimen2\font plus
\BIBentryALTinterwordstretchfactor\fontdimen3\font minus
  \fontdimen4\font\relax}
\providecommand\BIBforeignlanguage[2]{{%
\expandafter\ifx\csname l@#1\endcsname\relax
\typeout{** WARNING: IEEEtran.bst: No hyphenation pattern has been}%
\typeout{** loaded for the language `#1'. Using the pattern for}%
\typeout{** the default language instead.}%
\else
\language=\csname l@#1\endcsname
\fi
#2}}

\bibitem{ozawa}
R.~Ozawa and K.~Tahara, ``Grasp and dexterous manipulation of multi-fingered
  robotic hands: a review from a control view point,'' \emph{Advanced
  Robotics}, vol.~31, no. 19-20, pp. 1030--1050, 2017.

\bibitem{bicchi}
A.~Bicchi, ``Hands for dexterous manipulation and robust grasping: a difficult
  road toward simplicity,'' \emph{IEEE Transactions on Robotics and
  Automation}, vol.~16, no.~6, pp. 652--662, Dec 2000.

\bibitem{rahman}
N.~Rahman, L.~Carbonari, M.~D'Imperio, C.~Canali, D.~G. Caldwell, and
  F.~Cannella, ``A dexterous gripper for in-hand manipulation,'' in \emph{2016
  IEEE International Conference on Advanced Intelligent Mechatronics (AIM)},
  July 2016, pp. 377--382.

\bibitem{bircher_2}
W.~G. Bircher, A.~M. Dollar, and N.~Rojas, ``A two-fingered robot gripper with
  large object reorientation range,'' in \emph{2017 IEEE International
  Conference on Robotics and Automation (ICRA)}, May 2017, pp. 3453--3460.

\bibitem{chavan-dafle_2}
N.~C. Dafle, A.~Rodriguez, R.~Paolini, B.~Tang, S.~S. Srinivasa, M.~Erdmann,
  M.~T. Mason, I.~Lundberg, H.~Staab, and T.~Fuhlbrigge, ``Extrinsic dexterity:
  In-hand manipulation with external forces,'' in \emph{2014 IEEE International
  Conference on Robotics and Automation (ICRA)}, May 2014, pp. 1578--1585.

\bibitem{smith}
C.~Smith, Y.~Karayiannidis, L.~Nalpantidis, X.~Gratal, P.~Qi, D.~V.
  Dimarogonas, and D.~Kragic, ``Dual arm manipulation---a survey,''
  \emph{Robotics and Autonomous Systems}, vol.~60, no.~10, pp. 1340 -- 1353,
  2012.

\bibitem{park_1}
H.~A. Park and C.~S.~G. Lee, ``Extended cooperative task space for manipulation
  tasks of humanoid robots,'' in \emph{2015 IEEE International Conference on
  Robotics and Automation (ICRA)}, May 2015, pp. 6088--6093.

\bibitem{park_2}
------, ``Dual-arm coordinated-motion task specification and performance
  evaluation,'' in \emph{2016 IEEE/RSJ International Conference on Intelligent
  Robots and Systems (IROS)}, Oct 2016, pp. 929--936.

\bibitem{murooka}
M.~Murooka, Y.~Inagaki, R.~Ueda, S.~Nozawa, Y.~Kakiuchi, K.~Okada, and
  M.~Inaba, ``Whole-body holding manipulation by humanoid robot based on
  transition graph of object motion and contact,'' in \emph{2015 IEEE/RSJ
  International Conference on Intelligent Robots and Systems (IROS)}, Sept
  2015, pp. 3950--3955.

\bibitem{xian}
Z.~Xian, P.~Lertkultanon, and Q.~C. Pham, ``Closed-chain manipulation of large
  objects by multi-arm robotic systems,'' \emph{IEEE Robotics and Automation
  Letters}, vol.~2, no.~4, pp. 1832--1839, Oct 2017.

\bibitem{hang}
K.~Hang, M.~Li, J.~A. Stork, Y.~Bekiroglu, F.~T. Pokorny, A.~Billard, and
  D.~Kragic, ``Hierarchical fingertip space: A unified framework for grasp
  planning and in-hand grasp adaptation,'' \emph{IEEE Transactions on
  Robotics}, vol.~32, no.~4, pp. 960--972, Aug 2016.

\bibitem{sundaralingam}
B.~Sundaralingam and T.~Hermans, ``Relaxed-rigidity constraints: In-grasp
  manipulation using purely kinematic trajectory optimization,'' in
  \emph{Robotics: Science and Systems}, 2017.

\bibitem{psomopoulou}
E.~Psomopoulou, D.~Karashima, Z.~Doulgeri, and K.~Tahara, ``Stable pinching by
  controlling finger relative orientation of robotic fingers with rolling soft
  tips,'' \emph{Robotica}, vol.~36, no.~2, pp. 204--224, 2018.

\bibitem{chavan-dafle_4}
N.~Chavan-Dafle and A.~Rodriguez, ``Sampling-based planning of in-hand
  manipulation with external pushes,'' \emph{arXiv preprint arXiv:1707.00318},
  2017.

\bibitem{sintov_2}
A.~Sintov, O.~Tslil, and A.~Shapiro, ``Robotic swing-up regrasping manipulation
  based on the impulse-momentum approach and clqr control,'' \emph{IEEE
  Transactions on Robotics}, vol.~32, no.~5, pp. 1079--1090, Oct 2016.

\bibitem{cruciani}
S.~Cruciani and C.~Smith, ``In-hand manipulation using three-stages open loop
  pivoting,'' in \emph{2017 IEEE/RSJ International Conference on Intelligent
  Robots and Systems (IROS)}, Sept 2017, pp. 1244--1251.

\bibitem{shi_2}
J.~Shi, J.~Z. Woodruff, P.~B. Umbanhowar, and K.~M. Lynch, ``Dynamic in-hand
  sliding manipulation,'' \emph{IEEE Transactions on Robotics}, vol.~33, no.~4,
  pp. 778--795, Aug 2017.

\bibitem{papon}
J.~Papon, A.~Abramov, M.~Schoeler, and F.~Wörgötter, ``Voxel cloud
  connectivity segmentation - supervoxels for point clouds,'' in \emph{2013
  IEEE Conference on Computer Vision and Pattern Recognition}, June 2013, pp.
  2027--2034.

\end{thebibliography}

\end{document}